\documentclass[sigconf]{acmart}

\copyrightyear{2025}
\acmYear{2025}
\setcopyright{cc}
\setcctype[4.0]{by}
\acmConference[SAC '25]{The 40th ACM/SIGAPP Symposium on Applied Computing}{March 31-April 4, 2025}{Catania, Italy}
\acmBooktitle{The 40th ACM/SIGAPP Symposium on Applied Computing (SAC '25), March 31-April 4, 2025, Catania, Italy}
\acmDOI{10.1145/3672608.3707765}
\acmISBN{979-8-4007-0629-5/25/03}

\acmArticle{4}

\usepackage{savesym}
\savesymbol{Bbbk}
\savesymbol{openbox}

\usepackage{hyperref}
\usepackage[utf8]{inputenc}
\usepackage{graphicx}
\usepackage{amsmath}
\usepackage{amssymb}
\usepackage{amsthm}
\usepackage{MnSymbol}
\usepackage{booktabs}
\usepackage{algorithm}
\usepackage{algorithmic}
\restoresymbol{NEW}{Bbbk}
\restoresymbol{NEW}{openbox}

\usepackage{soul} 
\usepackage{verbatim} 

\usepackage{subcaption} 
\usepackage[justification=centering]{caption} 

\usepackage{multirow} 
\usepackage{listings} 

\usepackage[nameinlink]{cleveref} 
\usepackage{xurl} 
\usepackage{xspace} 

\emergencystretch=\maxdimen
\begin{document}
\title{Explainable Time Series Prediction of \\ Tyre Energy in Formula One Race Strategy}
\titlenote{© ACM 2025. This is the authors' version of the work. It is posted here for your personal use. Not for redistribution. The definitive Version of Record will be published in SAC 2025, \href{http://dx.doi.org/10.1145/3672608.3707765}{http://dx.doi.org/10.1145/3672608.3707765}.}


\renewcommand{\shorttitle}{SIG Proceedings Paper in LaTeX Format}

\author{Jamie Todd}
\affiliation{%
  \institution{Department of Computing,}
  \city{Imperial College London} 
  \country{UK}
}
\email{jamie.todd20@imperial.ac.uk}

\author{Junqi Jiang}
\affiliation{%
  \institution{Department of Computing,}
  \city{Imperial College London} 
  \country{UK}
}
\email{junqi.jiang20@imperial.ac.uk}

\author{Aaron Russo}
\affiliation{%
    \institution{Mercedes-AMG PETRONAS F1 Team}
    \city{Brackley}
    \country{UK}
}
\email{arusso@mercedesamgf1.com}

\author{Steffen Winkler}
\affiliation{%
    \institution{Mercedes-AMG PETRONAS F1 Team}
    \city{Brackley}
    \country{UK}
}
\email{swinkler@mercedesamgf1.com}

\author{Stuart Sale}
\affiliation{%
    \institution{Mercedes-AMG PETRONAS F1 Team}
    \city{Brackley}
    \country{UK}
}
\email{ssale@mercedesamgf1.com}

\author{Joseph McMillan}
\affiliation{%
    \institution{Mercedes-AMG PETRONAS F1 Team}
    \city{Brackley}
    \country{UK}
}
\email{jmcmillan@mercedesamgf1.com}

\author{Antonio Rago}
\authornote{Corresponding author}
\affiliation{%
  \institution{Department of Computing,}
  \city{Imperial College London} 
  \country{UK}
}
\email{a.rago@imperial.ac.uk}

\renewcommand{\shortauthors}{J. Todd et al.}

\begin{abstract}
Formula One (F1) race strategy takes place in a high-pressure and fast-paced environment where split-second decisions can drastically affect race results. Two of the core decisions of race strategy are when to make pit stops (i.e. replace the cars' tyres) and which tyre compounds (hard, medium or soft, in normal conditions) to select. The optimal pit stop decisions can be determined by estimating the tyre degradation of these compounds, which in turn can be computed from the energy applied to each tyre, i.e. the tyre energy. 
In this work, we trained deep learning models, using the Mercedes-AMG PETRONAS F1 team's historic race data consisting of telemetry, to forecast tyre energies during races. Additionally, we fitted XGBoost, a decision tree-based machine learning algorithm, to the same dataset and compared the results, with both giving impressive performance. 
Furthermore, we incorporated two different explainable AI methods, namely feature importance and counterfactual explanations, to gain insights into the reasoning behind the forecasts.
Our contributions thus result in an explainable, automated method which could assist F1 teams in optimising their race strategy.
\end{abstract}



\maketitle

\section{Introduction}
\label{sec:introduction}
Formula One (F1) is a popular sport all over the globe, with some Grand Prix races exceeding 100 million TV viewers\footnote{\href{https://www.formula1.com/en/latest/article/formula-1-announces-tv-race-attendance-and-digital-audience-figures-for-2021.1YDpVJIOHGNuok907sWcKW}{https://www.formula1.com/en/latest/article/formula-1-announces-tv-race-attendance-and-digital-audience-figures-for-2021.1YDpVJIOHGNuok907sWcKW}}. The significant prize pools mean that competing teams will go to great lengths to maximise the performance of their cars. Because of this, analysing historical race data is essential for teams as they provide useful insights into how various factors affect the car performance.
One of the most important factors is the tyre degradation, i.e. how different tyre compounds wear and degrade over time, which lowers the grip they provide and thus the car's speed. This means teams must optimise their strategies in their selection of which tyre compounds to use and when to pitstop, i.e. leave the race to change tyres.
A major contributor to the rate of tyre degradation is the fluctuating energy applied to each tyre, i.e. the tyre energy. This energy represents the sliding power, which is calculated using a physics-based model, comprising variables such as the forces on the tyre and the slip velocity.
AI models have not as of yet, as far as we are aware, been used in F1 to forecast tyre degradation. The industry standard for predicting this phenomenon is using simple linear models, which are clearly trustworthy but are often manually calculated and thus time-consuming, and do not take covariate data into account. 

Meanwhile, sophisticated \emph{time series prediction} models such as Long Short-Term Memory (LSTM) \cite{hochreiter} and Transformers \cite{vaswani} have been shown to accurately predict future unseen data when given historical multivariate data. 
It is therefore feasible that such a model could map car telemetry data such as speed and steering wheel angle into an estimation of the tyre energies at any given time. 
However, time series prediction models lack the ability to explain their forecasts to users. Given the high-risk nature of F1, race strategists would be highly unlikely to trust such an unaccountable model, since it is ultimately the strategist who would need to justify any decisions taken based on this information.
However, explainable AI (XAI) techniques have been shown to provide explanations of forecasts from time-series models (see \cite{rojat} for a recent survey), thus fostering trust from users towards the models.

In this work, we trained and optimised four different deep learning models, consisting of three different Recurrent Neural Network (RNN) models and a state-of-the-art Transformer-based model, to forecast\footnote{We use the terms describing the prediction and forecasting tasks interchangeably in the paper, since our outputs are the target variables at the next time step only.} tyre energies for an arbitrary race, given a set of covariate data, e.g. brake, throttle and steering wheel angle from the car's telemetry. In addition, we incorporated two baselines: linear regression and XGBoost, a gradient-boosted decision tree ensemble algorithm \cite{xgboost}. Furthermore, we tested alternative deployments of these prediction models such as single-track specialists and regression-based models to determine the optimal model for real-world use. Finally, we integrated two XAI algorithms, TIME \cite{sood} and CausalImpact \cite{brodersen}, to provide insights into the predictions of our deep learning models. While XGBoost was the most accurate at forecasting our test set, we find the results of our deep learning models to be promising. Furthermore, the integrated XAI methods offer rational and encouraging explanations for the predictions. Our work makes steps towards the real-world use of AI models in F1 for processing live data and predicting tyre energies during a race.

\section{Related Work}
\label{sec:related}
Time series prediction (also referred to as \textit{time series forecasting}) concerns training a model to predict the future values of \textit{targets} based on their previously observed values and/or that of \textit{covariates}. These covariates are external variables that can be used as inputs to help improve the accuracy of predictions.

Various deep learning architectures have been integrated into time series forecasting. RNNs have been the basis of many forecasting models, such as DeepAR by Salinas et al. \cite{salinas}. The LSTM architecture \cite{hochreiter}, a specialised form of RNN, has shown great potential in applying time series forecasting in fields such as macroeconomics \cite{kamolthip} and iron-making \cite{shockaert}. Transformers \cite{vaswani}, which were designed to handle sequence-to-sequence tasks, also see effective use in forecasting time series. PatchTST by Nie et al. \cite{nie}, a Transformer-based model, was the first to be trained on ``patched'' time series, where the input series is divided into equal-sized patches, inspiring other models such as TSMixer \cite{ekambaram}, a lightweight model composed entirely of multi-layer perception modules. 
Meanwhile, the Temporal Fusion Transformer (TFT) by Lim et al. \cite{lim} pushed the state-of-the-art in terms of both accuracy and innovation, consisting of an interpretability module, allowing insights such as feature importance to be analysed. Despite this, studies have shown that Transformer-based models may not be as effective at time series forecasting as previously thought, with simple baselines outperforming the state-of-the-art on benchmark datasets \cite{zeng}. 

With regards to applications of time series models in the world of motorsport, Schleinitz et al. \cite{schleinitz} combined a variational autoencoder \cite{kingma}, an anomaly detector and LSTM predictors to propose the variational autoencoder based selective prediction (VASP) framework. VASP was evaluated in the tasks of anomaly detection and robust time series prediction using data from motorsport race series. The framework outperformed the other approaches for this task and coped well even when anomalies were present. However, VASP, unlike our approach, was not deployed for time series prediction of tyre energy in F1.

Our work is not the first to apply machine learning models to predict variables concerning tyres.
Zhang et al. \cite{zhang} developed and trained a neural network to predict the tyre abrasion status in real-time. The tyre wear was predicted based on accelerometer and strain gauge data. All test data was predicted within $\pm0.5$mm, with 80\% of test cases within $\pm0.2$mm.
Meanwhile, Kim et al. \cite{kim} developed a 1D Convolutional Neural Network (CNN) that predicts tyre wear using various vehicle and tyre sensing information. Results showed that using more input data resulted in better performance overall, with the best models having a prediction error of 0.21mm, although models which only relied on wheel translation and rotation speed were still promising when accounting for sensing cost.

Accurate predictions alone may not warrant the use of such models in high-risk industries, as the models' lack of interpretability requires users to blindly trust their predictions, which would mean that risks are taken without justification.
XAI methods aim to help users understand how AI models compute outputs in an accessible manner \cite{bhatt}. The best XAI methods should be (among other features) interpretable and trustworthy \cite{rojat}, features which would undoubtedly be desirable in this application.

Feature importance is one of the most popular explanation types for tabular data, assigning an importance value to each feature reflecting how significant the feature was when computing the prediction \cite{bodria}. Temporal Importance Model Explanation (TIME) by Sood et al. \cite{sood} is an XAI method based on feature importance that calculates the importance of each input feature for the model at each time step. This is achieved via repeatedly perturbing the input tensor fed to the model and analysing the differences in results, enabling the user to know when and to what degree certain variables contribute to the model's output.

Counterfactual explanations assess how a model was dependent on external factors when making a particular prediction, focusing on the differences to obtain the opposite prediction \cite{bodria}. CausalImpact by Brodersen et al. \cite{brodersen} is a method based on counterfactual explanations which infers the impact of an ``intervention'' (an event that significantly affects future target values) on time series data, achieved by modelling the counterfactual scenario where the intervention did not occur and comparing the difference. 

In this paper, we will experiment with both TIME and CausalImpact.
To the best of our knowledge, there are no approaches in the literature which apply time series prediction or explainable AI to F1 applications, as we do in this paper.

\section{Time Series Forecasting of Tyre Energy}
\label{sec:setup}
In this section we detail our study, namely: 
the dataset used (§\ref{ssec:data});
the track state encodings (§\ref{ssec:encodings});
how the data was partitioned (§\ref{ssec:partition});
our problem formulation (§\ref{ssec:problem}); and 
the deployed forecasting models and XAI methods (§\ref{ssec:models}).

\subsection{Time Series Data}
\label{ssec:data}

We were provided with both serialised and tabular raw data by the Mercedes-AMG PETRONAS F1 Team, taken from real-time sources such as the car telemetry and GPS. After processing, aligning and merging the raw data, we produced a tabular time series of 0.1s resolution for each race and both drivers from the team, covering the 2020 to 2023 seasons. Some events were excluded as will be discussed in §\ref{sec: data_part}.
Each time series contained the following covariates:
\begin{itemize}
    \item Time into race (float);
    \item Car speed and its uncertainty (float);
    \item Lap number (int);
    \item Distance around track and distance uncertainty (float);
    \item Steering wheel angle (float);
    \item Under/oversteer coefficient (float);
    \item Lift and coast (boolean) and its distance (float);
    \item Gear (int); 
    \item Brake (percentage from 0-100);
    \item Throttle (percentage from 0-100);
    \item Engine speed (float);
    \item Fuel load (float);
    \item Track temperature (float);
    \item Pit stop status (boolean);
    \item DRS enabled (boolean) and Pod disabled (boolean);
    \item Track state (categorical, see §\ref{sec: track_state}).
\end{itemize}
The target variables were the energies applied to each of the four tyres. Time was measured from the race start (time = 0) until the end of the raw data sources (roughly five seconds after the car crosses the finish line). In addition to the above covariates, the car's latitude and longitude were stored for visualising model performance around the track, but were not an input to the model.

\subsection{Track State Encodings}
\label{sec: track_state}
\label{ssec:encodings}

The \textit{track state} is a categorical covariate, representing the current race conditions. In F1, \textit{green-flag} signifies standard conditions, \textit{yellow-flag} prohibits cars from overtaking due to a minor incident, and \textit{red-flag} conveys a race suspension, where cars must drive carefully to the pit lane. \textit{Safety cars} and \textit{virtual safety cars} deploy in response to track incidents, enforcing cars to adhere to strict speed limits. Thus, the track state significantly influences other covariates, such as car speed, and so four distinct encoding methods were evaluated:
\begin{itemize}
    \item \emph{exclude} the track state from the input data entirely;
    \item \emph{one-hot} encoding the state; 
    \item encode the state \emph{in order} based on expected car speed;
    \item filter the training set to include only time steps where the track state is \textit{green-flag}, then exclude the track state column.
\end{itemize}
Early tests gave no outright best method, so all approaches were included in hyperparameter optimisation experiments. 

\subsection{Dataset Partitioning}
\label{sec: data_part}
\label{ssec:partition}

Since there are a wide variety of tracks in F1, fixed training, validation and test sets were decided on with the following properties:
\begin{itemize}
    \item The training, validation and test sets consisted of 20, 6 and 7 events respectively, ranging from the 2020 to 2023 seasons. 
    \item Each event had 2 races, one for each Mercedes driver, with the end of one race being immediately followed by the start of the next race.
    \item The datasets contained a mixture of faster \emph{open} tracks and slower \emph{street} circuits, to explore the models' adaptability to various environments.
    \item The validation and test sets contained both \textit{seen} and \textit{unseen} tracks. A seen (unseen) track was one the models had (had not, respectively) been trained on from another season.
\end{itemize}
Races that took place under wet conditions, were suspended, or lacked necessary data were excluded from the datasets.

\subsection{Problem Formulation}
\label{ssec:problem}

For each time step of the test set, the time series forecasting model received the previous 100 time steps of input data and needed to forecast the immediate next value for the energies of each tyre (giving four targets in total). Contrary to standard forecasting tasks, the input data contained only covariate values; no past ground truth values for tyre energies were supplied to the model. The model's objective was to minimise the root mean squared error (RMSE) of the complete test set for the tyre energies.

\subsection{Forecasting Models and XAI Methods}
\label{ssec:models}

The following deep learning models were developed using the \textit{PyTorch} library\footnote{\href{https://pytorch.org/docs/stable/index.html}{https://pytorch.org/docs/stable/index.html}} for time series forecasting: RNN \cite{rumelhart}, LSTM \cite{hochreiter} and Gated Recurrent Unit (GRU) \cite{cho}. In addition, the TFT \cite{lim} was adapted for this task using the \textit{PyTorch Forecasting} library\footnote{\href{https://pytorch-forecasting.readthedocs.io/en/stable/index.html}{https://pytorch-forecasting.readthedocs.io/en/stable/index.html}}. 
Linear regression and XGBoost \cite{xgboost}, both statistical approaches, were integrated as baselines.

To provide insight into the models' predictions, we incorporated both TIME \cite{sood}, a temporal Feature Importance algorithm, and CausalImpact \cite{brodersen}, a statistical Counterfactual Explanation method.

\section{Evaluation and Discussion}
\label{sec:evaluation}
In this section, we evaluate and discuss the results of the time series forecasting models (§\ref{ssec:forecasting}) and the incorporated XAI algorithms (§\ref{ssec:XAI}). Our experiments were conducted on a Linux cluster with AMD EPYC 7742 CPU, 1TB RAM, and RTX 6000 GPU.

\subsection{Time Series Forecasting}
\label{ssec:forecasting}

We first performed hyperparameter tuning on the training set to optimise our deep learning models, utilising state-of-the-art algorithms provided by \textit{Ray Tune}\footnote{\href{https://docs.ray.io/en/latest/tune/index.html}{https://docs.ray.io/en/latest/tune/index.html}} and \textit{Optuna} \cite{optuna}. The track state encodings were included in the search space for each model, with the exception of the \emph{green-flag} only encoding, as this would be used as a comparison on the complete test set. The optimal configurations from these experiments are detailed in \Cref{tab: hyp_exp}.

\begin{table}
    \caption{Optimal Configurations from Hyperparameter Search Experiments}
    \centering
    \resizebox{\columnwidth}{!}{%
    \begin{tabular}{l|llll}
    \hline
        \textbf{Hyperparameter} & \textbf{RNN} & \textbf{LSTM} & \textbf{GRU} & \textbf{TFT} \\ \hline
        \textbf{Track State Encoding} & One-Hot & Exclude & One-Hot & One-Hot \\ 
        \textbf{Gradient Clip Norm} & 0.721 & 0.375 & 0.124 & 0.811 \\ 
        \textbf{Hidden Size} & 256 & 512 & 256 & 32 \\ 
        \textbf{Learning Rate} & 0.073 & 0.002 & 1e-4 & 4e-5 \\ 
        \textbf{Number of Layers} & 8 & 8 & 4 & 8 \\ 
        \textbf{Optimiser} & SGD & Ranger & AdamW & Adam \\ 
        \textbf{Bi-directional} & NO & YES & NO & - \\ 
        \textbf{Dropout} & - & - & - & 0.150 \\ 
        \textbf{\# Attention Heads} & - & - & - & 8 \\ \hline
    \end{tabular}}
    \label{tab: hyp_exp}
\end{table}

The results for the test set are presented in \Cref{tab: final_results}. \textbf{HS Best} represents the track state encoding featured in the best configuration from hyperparameter search experiments (from Table \ref{tab: hyp_exp}), while \textbf{Green-flag} represents an identical model with a green-flag encoded dataset. Values in \textit{italics} highlight the lowest RMSE of the two track state encodings for each model, with the lowest overall highlighted in bold.

\begin{table}
    \caption{Baseline and Fine-tuned Deep Learning Model RMSEs on Test Set}
    \centering
    \resizebox{\columnwidth}{!}{%
    \begin{tabular}{ccccccc} \hline
        \textbf{Model} & \textbf{Linear} & \textbf{XGBoost} & \textbf{RNN} & \textbf{LSTM} & \textbf{GRU} & \textbf{TFT} \\ \hline
        \textbf{HS Best} & \textit{7.893} & 4.841 & 5.926 & 5.896 & \textit{5.517} & 5.693 \\ 
        \textbf{Green-flag} & 7.916 & \textbf{4.785} & \textit{5.697} & \textit{5.730} & 5.558 & \textit{5.305} \\ \hline
    \end{tabular}}
    \label{tab: final_results}
\end{table}

From the results, one can observe that all neural models vastly outperformed the linear regression baseline, with the TFT producing the most accurate forecasts of the deep learning models. XGBoost outperformed all trained deep learning models by a considerable margin. However, classifying XGBoost as a baseline for this task may be somewhat unfair as it had to fit to the complete training set and thus required far more RAM (of up to 1024 GB) than any of our deep learning models (at most 256 GB).

\begin{figure*}
    \centering
    \begin{subfigure}[b]{0.49\linewidth}
        \centering
        \includegraphics[width=\linewidth]{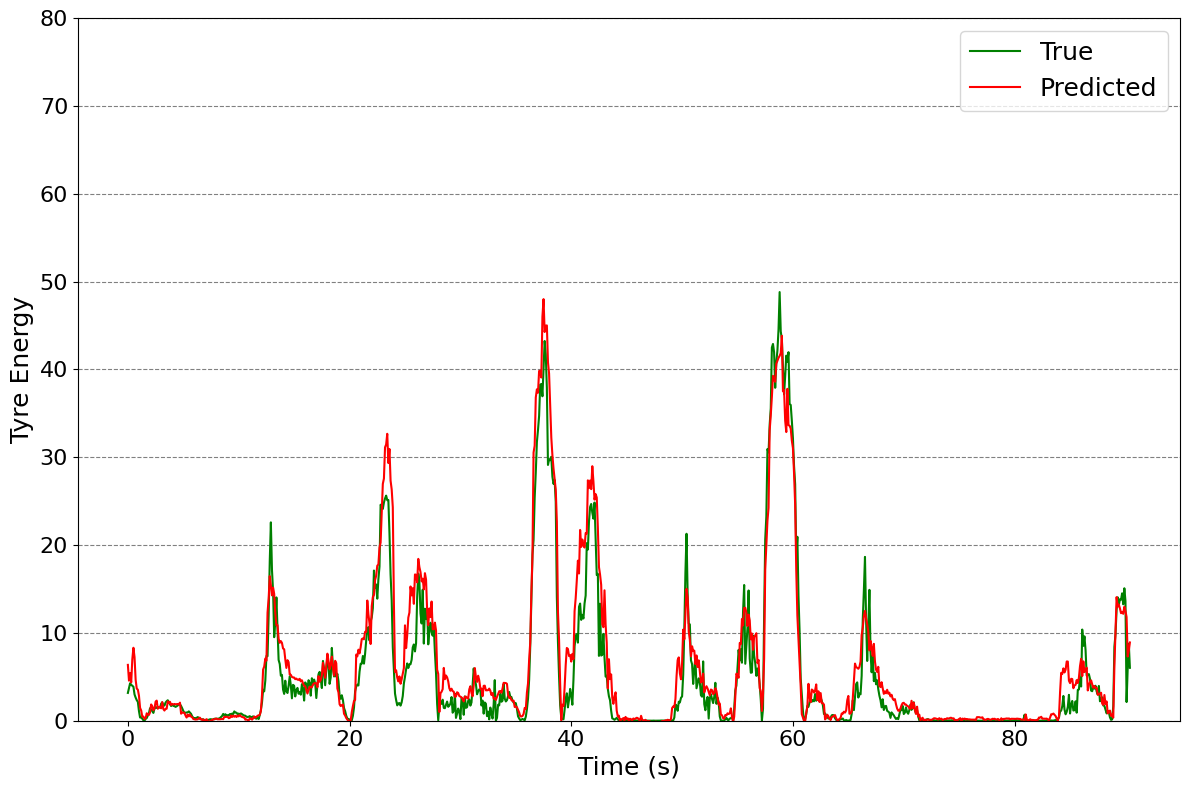}
    \end{subfigure}
    \hfill
    \begin{subfigure}[b]{0.49\linewidth}
        \centering
        \includegraphics[width=\linewidth]{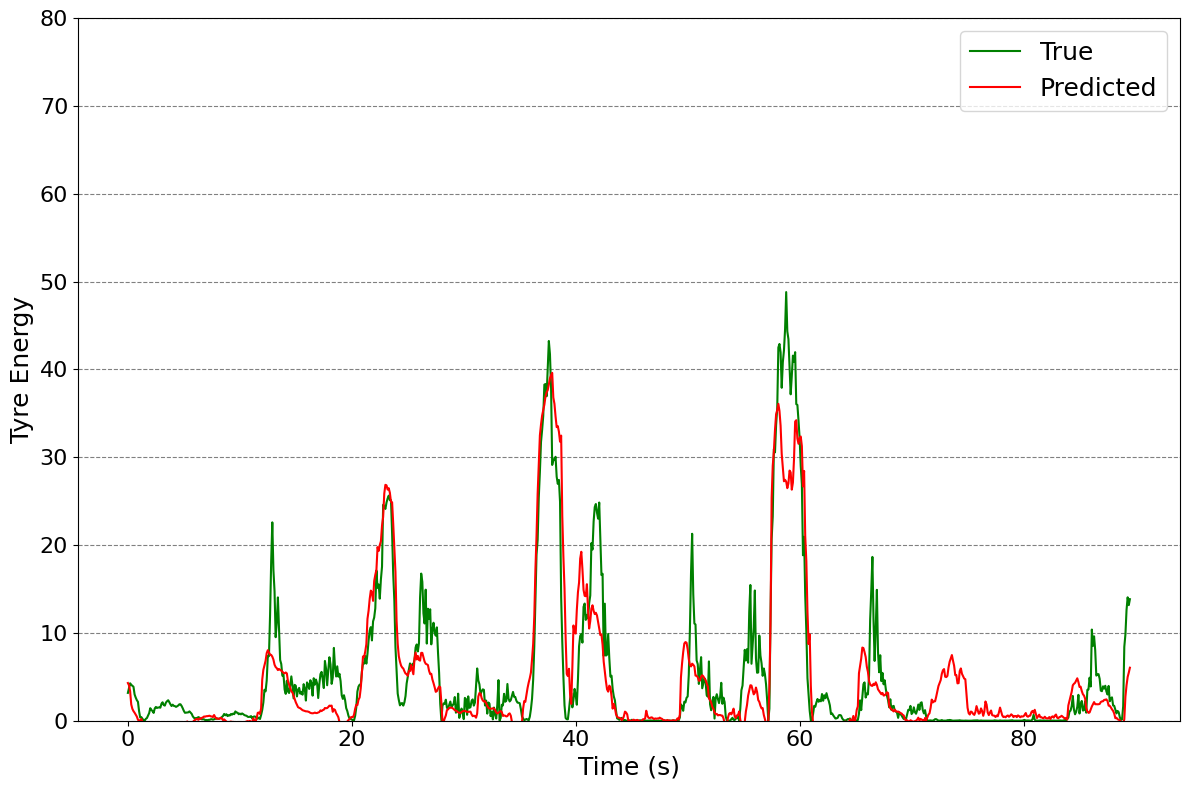}
    \end{subfigure}
    \caption{Green-flag XGBoost \textit{(left)} and TFT \textit{(right)} front-left tyre energy forecasts for one lap of a seen open track}
    \label{fig: abu_graph}
\end{figure*}

\begin{figure*}
    \centering
    \begin{subfigure}[b]{0.49\linewidth}
        \centering
        \includegraphics[width=\linewidth]{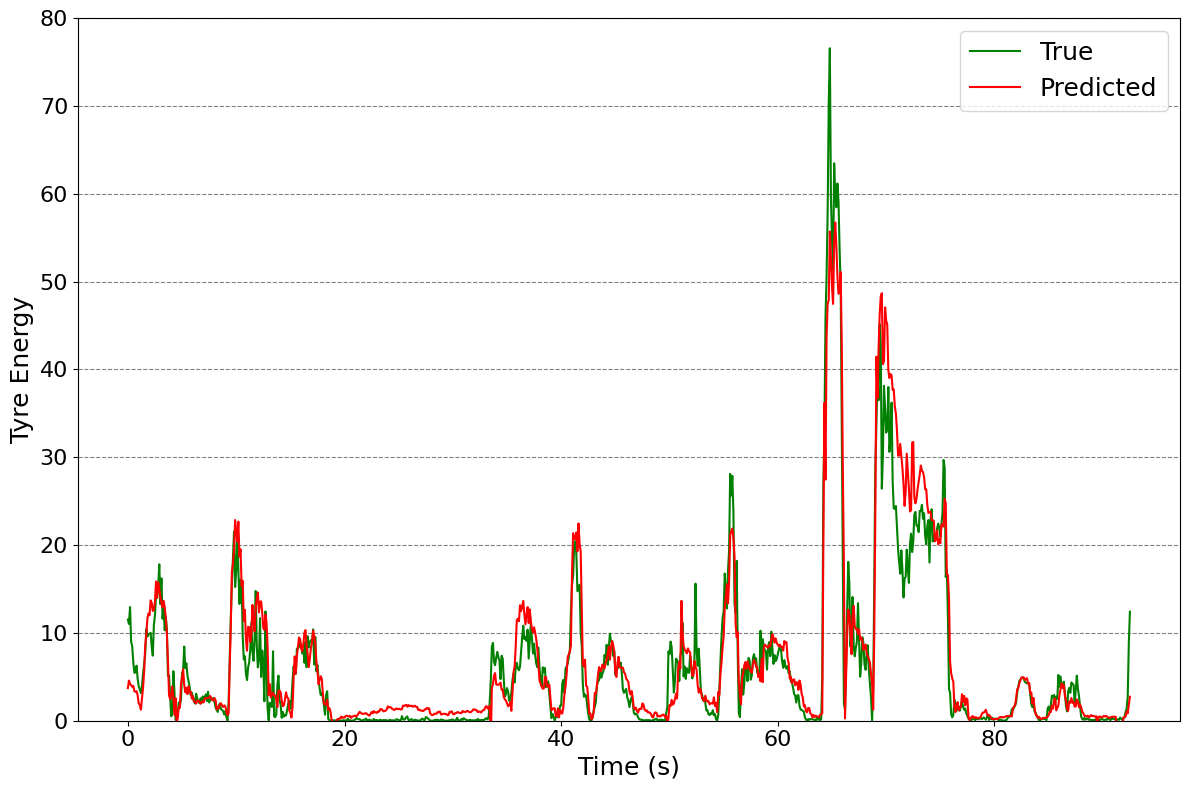}
    \end{subfigure}
    \hfill
    \begin{subfigure}[b]{0.49\linewidth}
        \centering
        \includegraphics[width=\linewidth]{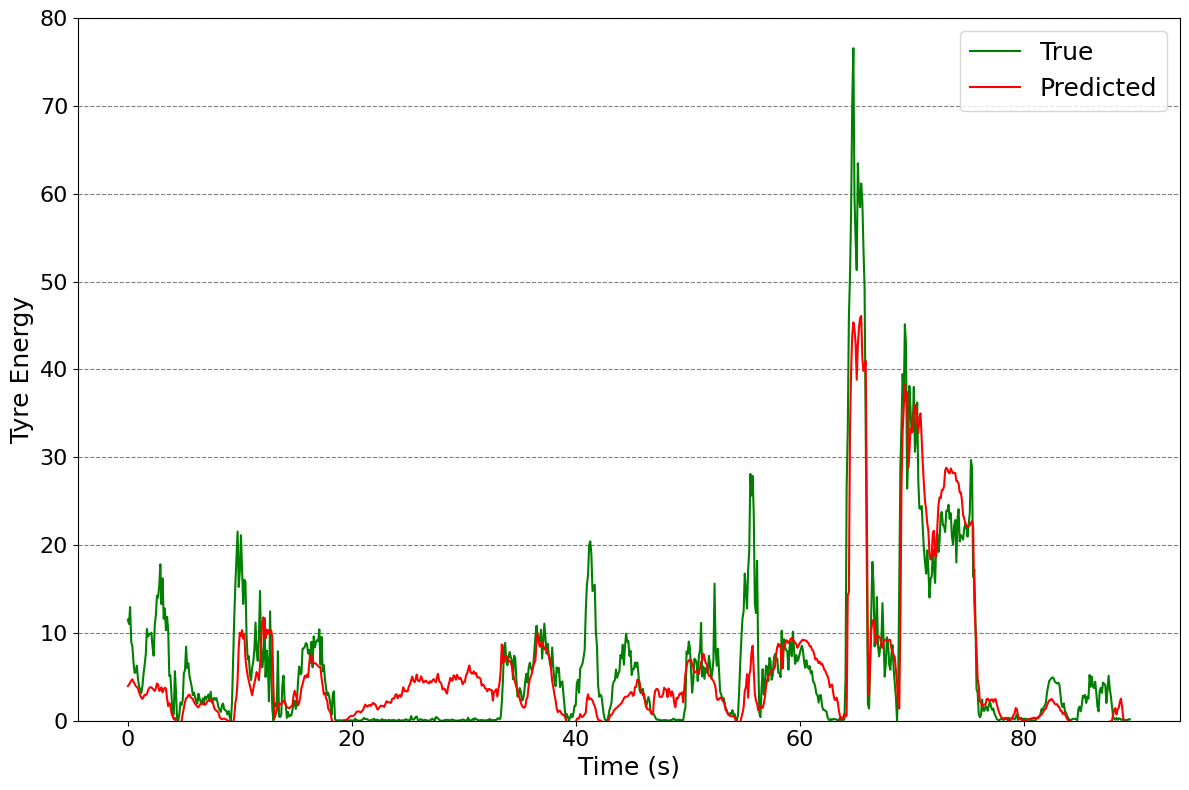}
    \end{subfigure}
    \caption{Green-flag XGBoost \textit{(left)} and TFT \textit{(right)} front-right tyre energy forecasts for one lap of an unseen street circuit}
    \label{fig: miami_graph}
\end{figure*}

Regardless, XGBoost's performance was very promising, as shown in \Cref{fig: abu_graph}, which also displays the TFT's forecasts over the same race and lap for comparison. XGBoost captures the trends of the lap far more closely than the TFT. \Cref{fig: miami_graph} further demonstrates this on an unseen street circuit, showcasing XGBoost's adaptability to new (types of) tracks.

\begin{figure*}
    \centering
    \begin{subfigure}[b]{0.41\linewidth}
        \centering
        \includegraphics[width=\linewidth]{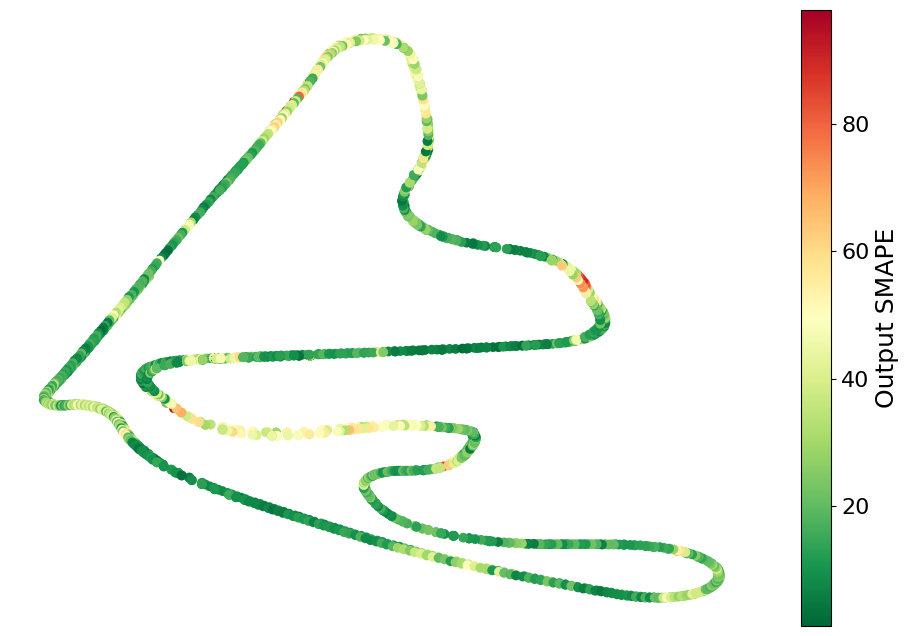}
    \end{subfigure}
    \hfill
    \begin{subfigure}[b]{0.41\linewidth}
        \centering
        \includegraphics[width=\linewidth]{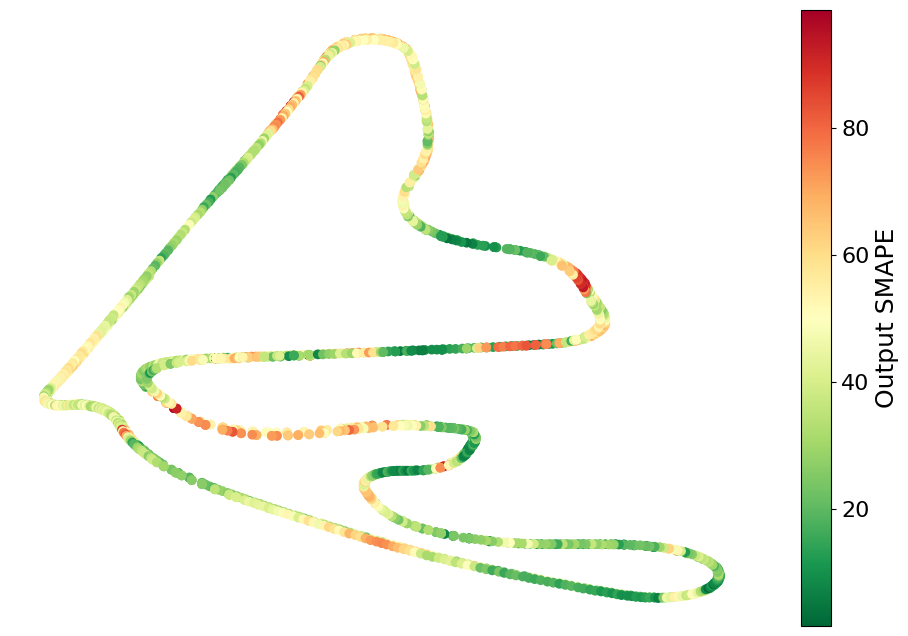}
    \end{subfigure}
    \caption{XGBoost \textit{(left)} and TFT \textit{(right)} SMAPE values over several laps of the Abu Dhabi Grand Prix (a seen open track)}
    \label{fig: abu_track}
\end{figure*}

\begin{figure*}
    \centering
    \begin{subfigure}[b]{0.41\linewidth}
        \centering
        \includegraphics[width=\linewidth]{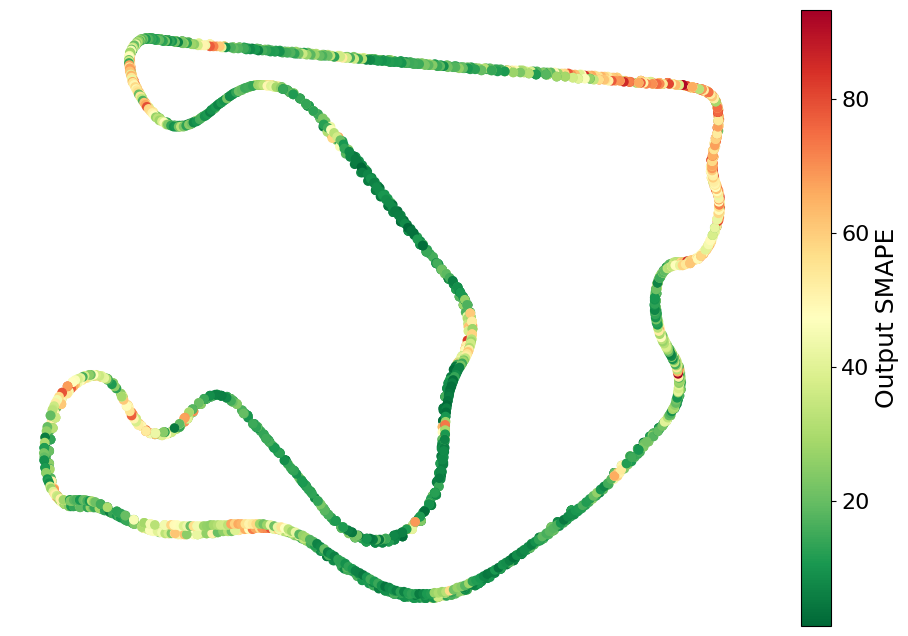}
    \end{subfigure}
    \hfill
    \begin{subfigure}[b]{0.41\linewidth}
        \centering
        \includegraphics[width=\linewidth]{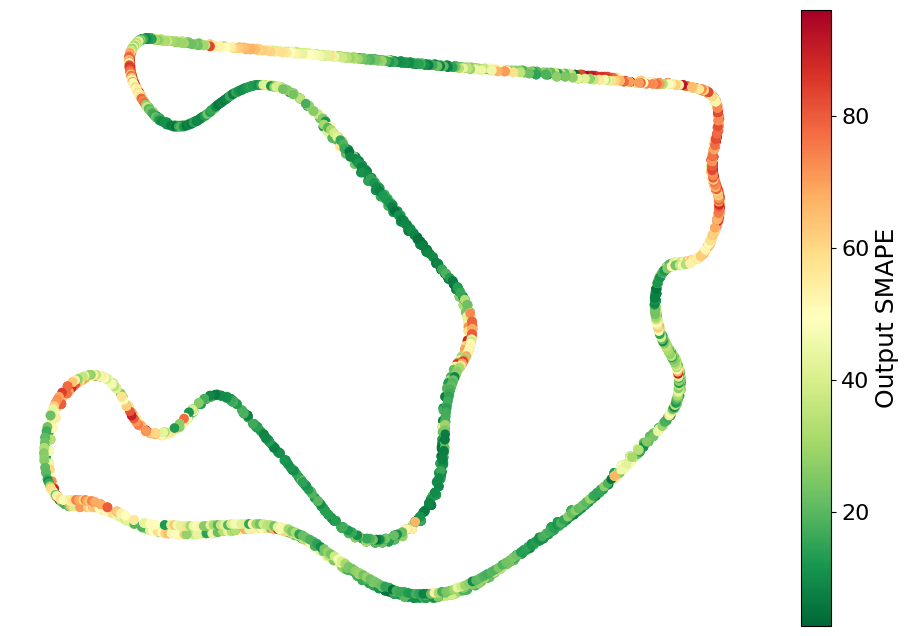}
    \end{subfigure}
    \caption{XGBoost \textit{(left)} and TFT \textit{(right)} SMAPE values over several laps of the Miami Grand Prix (an unseen street circuit)}
    \label{fig: miami_track}
\end{figure*}

Figures \ref{fig: abu_track} and \ref{fig: miami_track} demonstrate the performance of XGBoost and TFT over the Abu Dhabi Grand Prix, a seen open track, and the Miami Grand Prix, an unseen street circuit, respectively. These plots highlight the symmetric mean absolute percentage error (SMAPE), a proportional error metric ranging from 0 to 100, allowing us to determine the strongest and weakest sections of the tracks. XGBoost has far fewer sections of the Abu Dhabi track with high SMAPE compared to the TFT, although the difference between the two is less significant for the Miami track. Most weak sections are either during, or shortly preceding or following a corner, which is reasonable as tyre energies change rapidly during these moments.

\begin{figure*}
    \centering
    \includegraphics[width=1\linewidth]{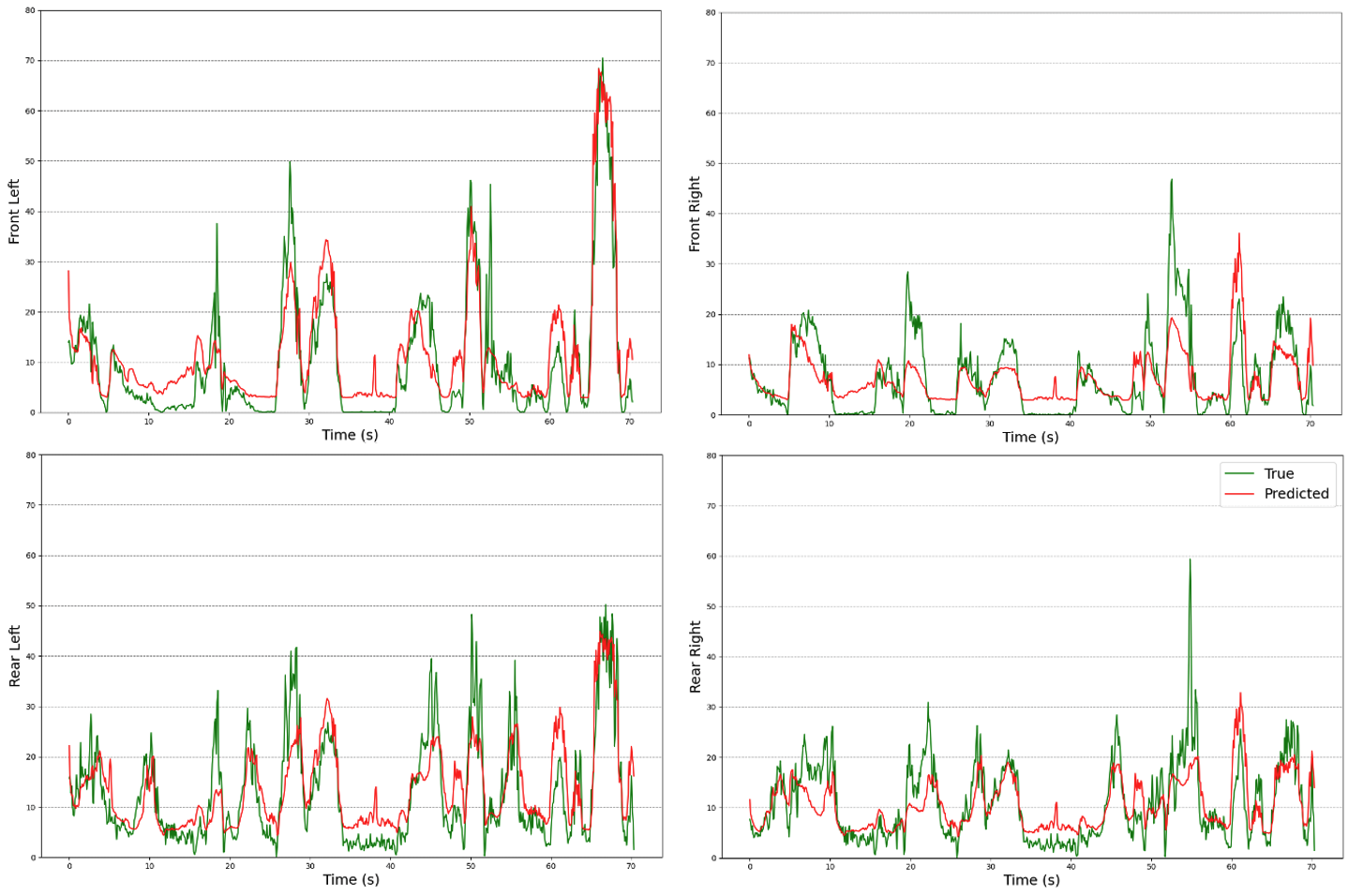}
    \caption{Green-flag TFT all tyre energy forecasts for one lap of an unseen \emph{clockwise} open circuit.\\The top row displays the front left and right tyres, while the bottom row displays the rear left and right tyres.}
    \label{fig: tft_zand}
\end{figure*}

\Cref{fig: tft_zand} details the green-flag TFT forecasts for tyres over one lap of an unseen open track. The cars follow this track clockwise, meaning it predominantly consists of right turns, and this is demonstrated by the larger energy spikes on average for the front left (compared to the front right) and rear left (compared to the rear right) tyres. This further demonstrates that our models are replicating real-world phenomena exhibited by the cars' dynamics.

\Cref{tab: seen_unseen_races} demonstrates the results for the GRU and TFT models on seen and unseen tracks respectively. It is clear that the models perform better on seen tracks, as would be expected, but the performance difference is more obvious for the GRU models. Moreover, GRU outperforms TFT for seen races, whereas the green-flag TFT is significantly stronger for unseen events, showcasing the adaptability of the TFT.

\begin{table*}
    \caption{GRU and TFT models evaluated on several \textit{seen} (Abu Dhabi, Bahrain, Austria, Monaco and Spain) and \textit{unseen} (Netherlands, Qatar, Hungary, Miami and Azerbaijan) tracks, with mean results}
    \centering
    \resizebox{2.1\columnwidth}{!}{%
    \begin{tabular}{lcccccccccccc}
    \cline{2-13}
         & \multicolumn{5}{c}{\textbf{Seen Tracks}} & \textbf{Seen} & \multicolumn{5}{c}{\textbf{Unseen Tracks}} & \textbf{Unseen} \\
    \cline{1-6}
    \cline{8-12}
        \textbf{Model} & \textbf{Abu D.} & \textbf{Bahrain} & \textbf{Austria} & \textbf{Monaco} & \textbf{Spain} & \textbf{Mean} & \textbf{Neth.} & \textbf{Qatar} & \textbf{Hungary} & \textbf{Miami} & \textbf{Azer.} & \textbf{Mean} \\ \hline
        \textbf{One-hot GRU} & 6.504 & \textbf{4.218} & \textbf{5.922} & \textbf{4.595} & \textbf{5.164} & \textbf{5.281} & 5.980 & 5.812 & 7.180 & 5.696 & 5.094 & 5.952 \\ 
        \textbf{One-hot TFT} & 7.154 & 4.508 & 6.854 & 4.760 & 5.600 & 5.775 & 6.140 & 6.766 & 6.633 & 5.680 & \textbf{4.644} & 5.972 \\ 
        \textbf{Green-flag GRU} & \textbf{6.393} & 4.279 & 6.257 & 4.753 & 5.472 & 5.431 & \textbf{4.817} & 5.949 & 7.102 & 5.500 & 5.309 & 5.735 \\ 
        \textbf{Green-flag TFT} & 6.666 & 4.188 & 6.336 & 4.713 & 5.355 & 5.452 & 5.171 & \textbf{5.739} & \textbf{6.360} & \textbf{5.260} & 4.843 & \textbf{5.475} \\ \hline
    \end{tabular}}
    \label{tab: seen_unseen_races}
\end{table*}

The TFT models were also compared against \textit{single-track specialist} (STS) TFT models, which were trained on one track over multiple seasons and were evaluated on the same track for a more recent season. We speculate that, while STS models do not generalise well for arbitrary tracks, their over-fitting to the circuit would be useful in real-world applications, e.g. for planning the race strategy for a single upcoming race. \Cref{tab: sts} demonstrates that STS models tend to improve upon more generalised models for their specialised track, supporting this hypothesis. Monaco is an exception to this, but unlike the other evaluated tracks, the STS model trained on a single Monaco race only, which suggests that the issue here may have been a lack of data. It should also be noted that Monaco has the lowest tyre degradation, and thus tyre energies, of all tracks, and is often regarded as an exception.\footnote{\href{https://www.formula1.com/en/latest/article/what-tyres-will-the-teams-and-drivers-have-for-the-2022-monaco-grand-prix.15S0GNtRiZBTCaOT2T7zUO}{https://www.formula1.com/en/latest/article/what-tyres-will-the-teams-and-drivers-have-for-the-2022-monaco-grand-prix.15S0GNtRiZBTCaOT2T7zUO}}

\begin{table}
    \caption{TFT models compared against STS TFT models for three distinct tracks, as well as the mean}
    \centering
    \resizebox{\columnwidth}{!}{%
    \begin{tabular}{lcccc} \hline
        \textbf{Model} & \textbf{Abu D.} & \textbf{Austria} & \textbf{Monaco} & \textbf{Mean} \\ \hline
        \textbf{One-hot TFT} & 7.154 & 6.854 & 4.760 & 6.256 \\ 
        \textbf{Green-flag TFT} & 6.666 & 6.336 & \textbf{4.713} & 5.905 \\ 
        \textbf{STS One-hot TFT} & \textbf{5.535} & \textbf{4.849} & 5.115 & \textbf{5.166} \\ 
        \textbf{STS Green-flag TFT} & 5.834 & 5.300 & 5.269 & 5.468 \\ \hline
    \end{tabular}}
    \label{tab: sts}
\end{table}

\begin{table}
    \caption{\underline{For}ecasting TFT compared against \underline{Reg}ression TFTs for \underline{G}reen-flag and \underline{E}xclude encodings}
    \centering
    \resizebox{\columnwidth}{!}{%
    \begin{tabular}{lccccc}
    \hline
        \textbf{Model} & \textbf{Azer.} & \textbf{Spain} & \textbf{Miami} & \textbf{Neth.} & \textbf{Test Set} \\ \hline
        \textbf{For. G TFT} & 4.843 & \textbf{5.355} & \textbf{5.260} & \textbf{5.171} & \textbf{5.305} \\ 
        \textbf{Reg. E TFT} & \textbf{4.572} & 6.163 & 6.131 & 5.393 & 5.558 \\ 
        \textbf{Reg. G TFT} & 4.676 & 6.183 & 6.034 & 5.229 & 5.877 \\ \hline
    \end{tabular}}
    \label{tab: regression}
\end{table}

As a final investigation, regression-based TFT models were trained and evaluated on the complete test set as well as individual tracks. This was to test the hypothesis that 100 time steps of covariate data may be too much information for the models and thus harm the performance. These regression-based models only received one time step of covariate data and must predict the tyre energies for that same time step. From the results in \Cref{tab: regression}, we found the TFT to be more accurate than its regression-based counterpart, although the difference in performance was not significant given the large disparity in input data size.
\subsection{Explainable AI}
\label{ssec:XAI}

\begin{figure}
    \centering
    \includegraphics[width=1\linewidth]{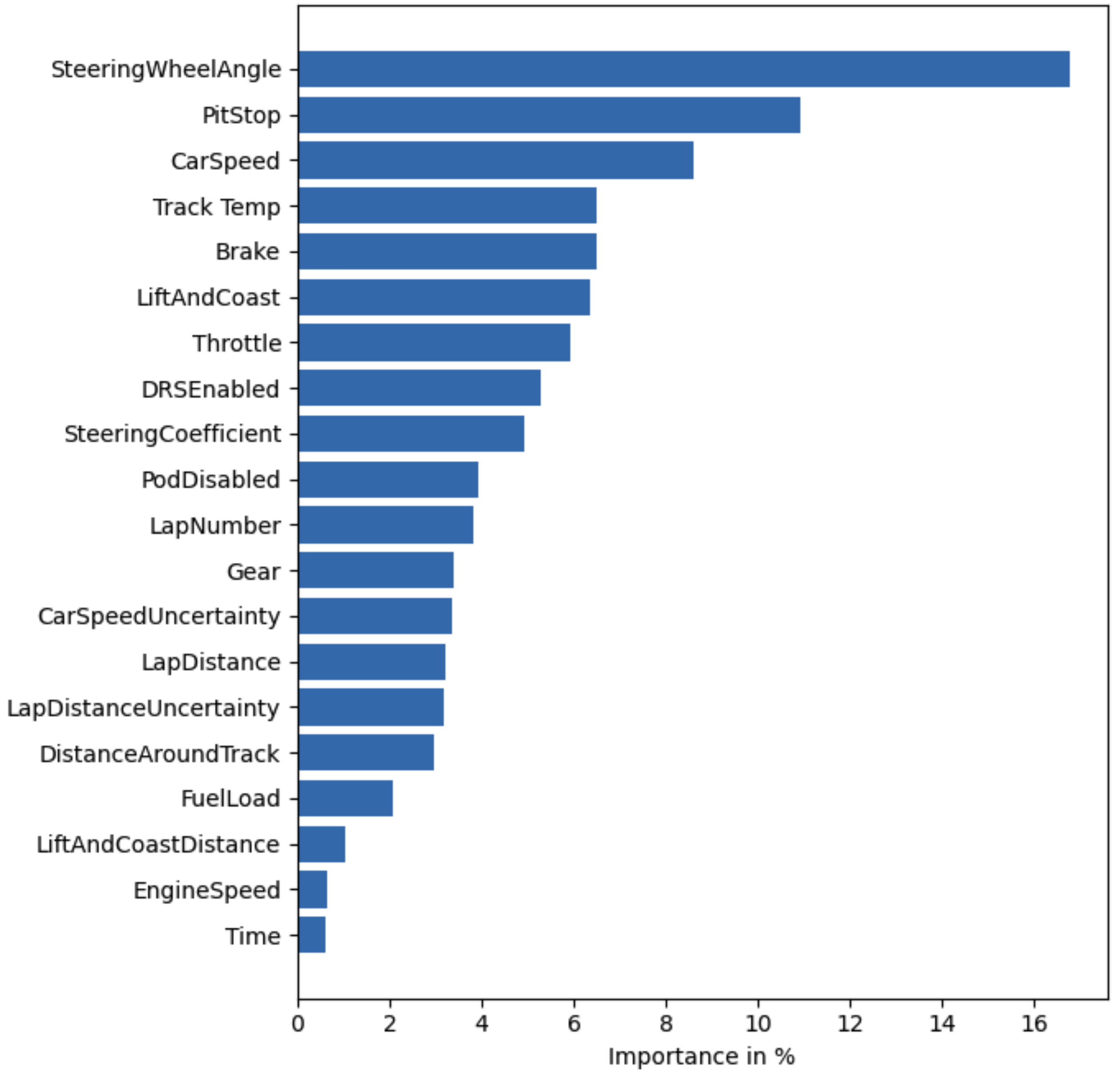}
    \caption{In-built TFT feature importance}
    \label{fig: tft_fi_enc}
\end{figure}

\Cref{fig: tft_fi_enc} ranks the feature importance of the observed covariates for the green-flag TFT, taken from its in-built interpretability module. \emph{SteeringWheelAngle}, which affects how sharply the car takes corners, is the most important variable for forecasting tyre energies. Overall, the ranking of the variables is logical and is along the lines of what we were expecting, with the most relevant covariates having greater significance. For example, \textit{PitStop}, which signifies whether a pit stop is taking place, is the second most important variable. This is intuitive as a car is stationary during a pit stop and thus no energy is applied to the tyres.

\begin{figure*}
    \centering
    \includegraphics[width=1\linewidth]{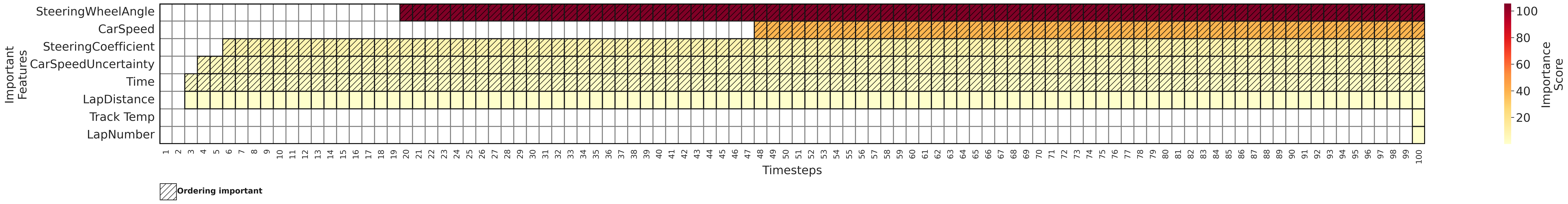}
    \caption{TIME heat map of green-flag TFT for Miami, front-left tyre}
    \label{fig: time_tft_mia_fl}
\end{figure*}

\begin{figure*}
    \centering
    \includegraphics[width=1\linewidth]{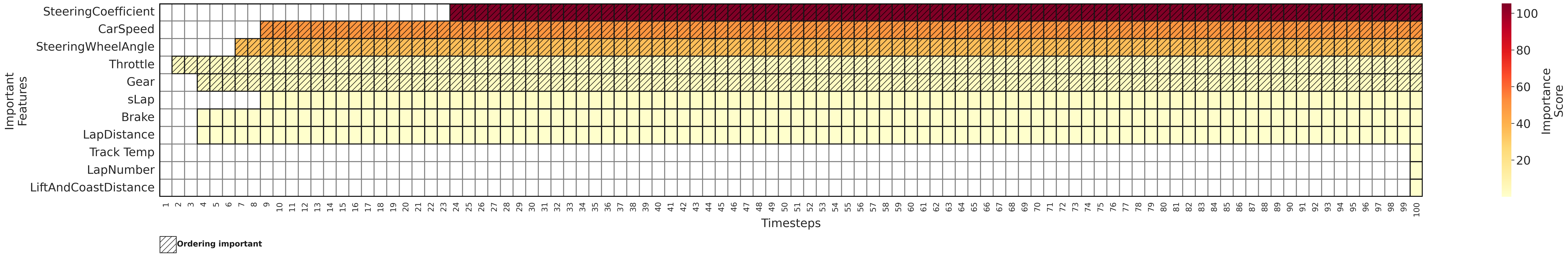}
    \caption{TIME heat map of one-hot STS for Abu Dhabi, front-left tyre}
    \label{fig: time_sts_abu_fl}
\end{figure*}

TIME \cite{sood} produces a heat map as shown in \Cref{fig: time_tft_mia_fl}, which displays the temporal feature importance for the generalised TFT model. This heat map covers a ten-second window of a corner turn, including the entrance, apex and exit.
Covariates corresponding to the car speed, steering wheel angle and steering coefficient are assigned high importance for the model. Steering wheel angle is only considered significant when the car starts turning; similarly, car speed is only relevant after the apex when the car starts to accelerate. This is in line with expectations, with other models producing similar results (see \Cref{fig: time_sts_abu_fl}).

\begin{figure}
    \centering
    \includegraphics[width=1\linewidth]{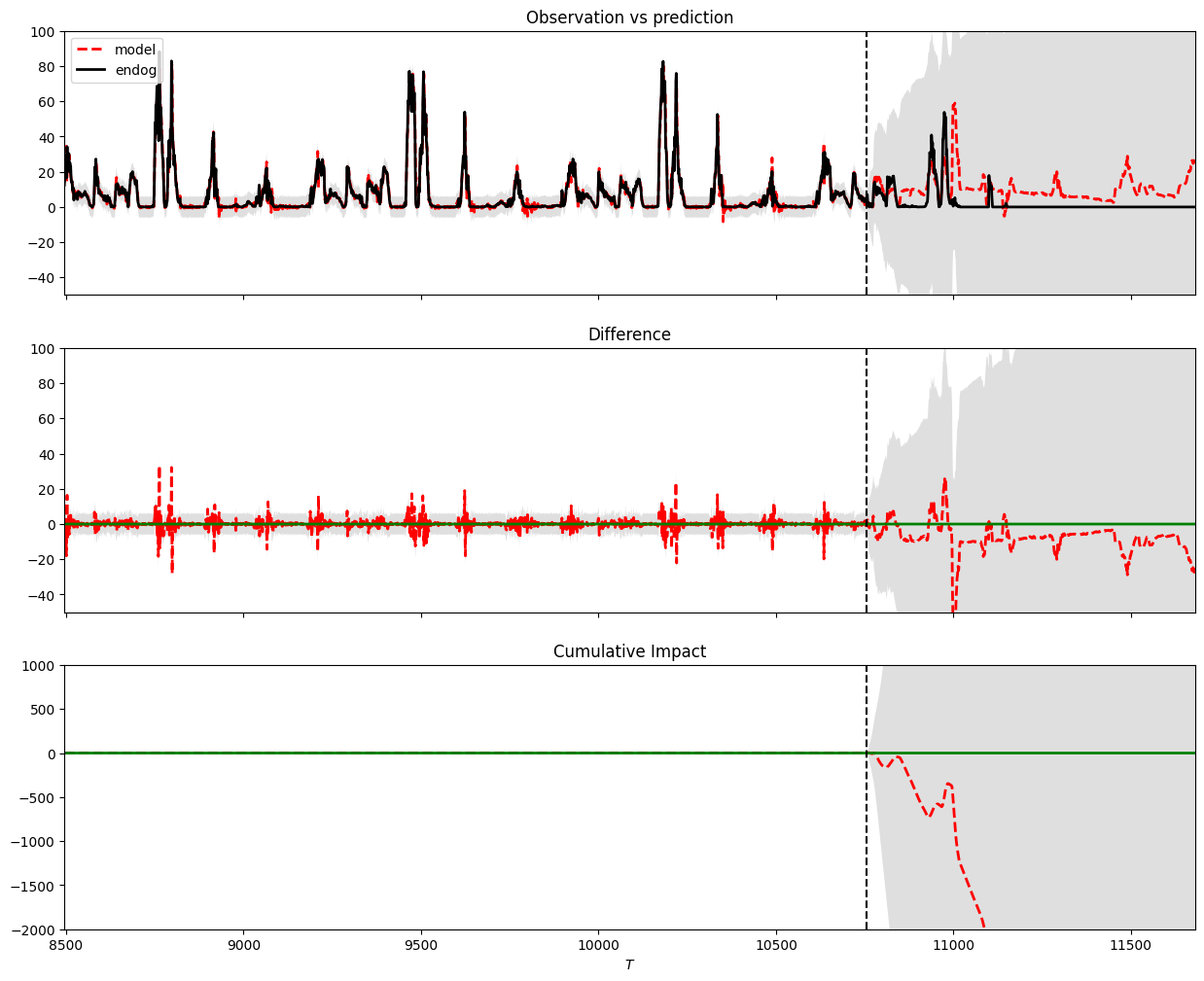}
    \caption{Causal impact of front-left tyre energies for a virtual safety car incident. The first graph gives the counterfactual predictions if the virtual safety car was not deployed (red dashed line), compared to the ground truth observations (black solid line). The second plot displays the difference between the two and the third demonstrates the cumulative difference.}
    \label{fig: ci_a1r}
\end{figure}

CausalImpact \cite{brodersen} was utilised to analyse the impact of an incident during a race which led to the virtual safety car being deployed. The results are demonstrated in \Cref{fig: ci_a1r}. The counterfactual is the hypothetical scenario where the virtual safety car was not deployed i.e., if the race continued under green-flag conditions. The grey shaded area represents the 95\% confidence interval, which is wide given the volatility of the tyre energies. The true energies were very low during this period, but the predictions reflect a typical lap. Thus, the difference is almost always negative and the cumulative impact decreases rapidly. The algorithm also calculates the probability of causal effect at 57.5\%. This demonstrates that the virtual safety car had a significant impact on reducing the tyre energies during its active period.

\section{Conclusions and Future Work}
\label{sec:conclusions}
In this work, we implemented and trained three different recurrent neural network models, and integrated a state-of-the-art transformer model, to forecast tyre energies given only past covariate data. While all of these models vastly outperformed the linear regression baseline, XGBoost proved to be a very powerful tool for this particular time series forecasting task, albeit with its weaknesses such as high memory requirements and rigid training time. Despite this, we found alternative methods to improve model accuracy, such as specialising models to particular tracks. Furthermore, we incorporated XAI techniques to justify the models' predictions.

Future work will be focused on extending the models' capabilities to handle practice and qualifying session data to boost accuracy, as well as reconsidering which covariates could be added or removed to improve performance. More explainability methods could be utilised and assessed alongside those we have introduced here, both empirically with datasets and in user studies with race strategists to determine their effectiveness along typical XAI properties such as \emph{faithfulness} \cite{Jacovi_20}, i.e. how closely the explanations replicate the models, and \emph{trust} \cite{Jacovi_21}, i.e. how much users are willing to trust the model. The real-world use of our models is currently limited to post-race analysis due to their iterative processing of complete race data. Enhancing our models to process real-time data and compute predictions on the fly will extend their use to live races, potentially giving F1 teams an edge over their competitors. Finally, we would like to investigate how our models translate to the wider automotive industry, given the increasing importance of tyre degradation for air pollution as electric vehicles become widespread \cite{Katsikouli_20}.

\bibliographystyle{ACM-Reference-Format}
\bibliography{bib}

\end{document}